\documentclass{article}

\PassOptionsToPackage{numbers, compress}{natbib}

\usepackage[final]{nips_2018}

\usepackage[utf8]{inputenc} 
\usepackage[T1]{fontenc}    
\usepackage{hyperref}       
\usepackage{url}            
\usepackage{booktabs}       
\usepackage{amsfonts}       
\usepackage{nicefrac}       
\usepackage{microtype}      
\usepackage{graphicx}
\usepackage{comment}
\usepackage{soul}
\usepackage{xcolor}

\title{Disease Detection in Weakly Annotated Volumetric Medical Images using a Convolutional LSTM Network}

\author{
  Nathaniel Braman \\
  Department of Biomedical Engineering\\
  Case Western Reserve University\\
  Cleveland, OH 44106\\
  \texttt{nathaniel.braman@case.edu} \\
   \And
   David Beymer \\
   IBM Almaden Research Center \\
   San Jose, CA 95120\\
   \texttt{beymer@us.ibm.com } \\
   \And
   Ehsan Dehghan \\
   IBM Almaden Research Center \\
   San Jose, CA 95120 \\
   \texttt{edehgha@us.ibm.com} \\
}

\begin{document}

\maketitle

\begin{abstract}
We explore a solution for learning disease signatures from weakly, yet easily obtainable, annotated volumetric medical imaging data by analyzing 3D volumes as a sequence of 2D images. We demonstrate the performance of our solution in the detection of emphysema in lung cancer screening low-dose CT images. Our approach utilizes convolutional long short-term memory (LSTM) to “scan” sequentially through an imaging volume for the presence of disease in a portion of scanned region. This framework allowed effective learning given only volumetric images and binary disease labels, thus enabling training from a large dataset of 6,631 un-annotated image volumes from 4,486 patients. When evaluated in a testing set of 2,163 volumes from 2,163 patients, our model distinguished emphysema with area under the receiver operating characteristic curve (AUC) of .83. This approach was found to outperform 2D convolutional neural networks (CNN) implemented with various multiple-instance learning schemes (AUC=0.69-0.76) and a 3D CNN (AUC=.77). 
\end{abstract}

\section{Introduction}
\vspace{-4mm}

 One of the most significant obstacles to the development of deep learning-based computer-aided diagnosis (CAD) platforms in radiology is the need for large, annotated medical image datasets. Particularly in the case of 3D imaging modalities, such as computed tomography (CT), it is often prohibitively onerous for radiologists to provide sufficient manual annotations for the training of deep models. Therefore, training a model using a large data set of annotated samples is practically unfeasible. One such domain is the detection of emphysema, a disease associated with shortness of breath and elevated cancer risk. Emphysema often manifests as ruptured air sacs within only a portion of the lung volume. Its variety of presentations on CT presents a challenge to training a model to detect emphysema from volumetric imaging data with binary diagnostic labels alone. 

A commonly utilized approach to enable learning with the absence of precise labels is multiple instance learning (MIL). In MIL, sets of samples are grouped into labeled bags, wherein a positive label signifies the presence of positive samples within a bag. Previous work has successfully leveraged an MIL framework for detection of emphysema and a variety of other lung diseases on CT. MIL using a hand-crafted feature-based classifier to evaluate a number of 2D patches from the lung has been shown to identify emphysema \cite{emph1,emph2} and other lung diseases \cite{copd}. More recently, Bortsova et al. \cite{miccai} reported success in grading emphysema by summarizing the output of a convolutional neural network (CNN) over a number of 2D patches using a proportional approach similar to MIL. 

A disadvantage of MIL-based approaches is that they fail to retain relationships between samples. For example, while effective at summarizing information from a number of samples, MIL does not retain the spatial relationship between samples drawn from an image. Furthermore, the efficacy of MIL is dependent on the pooling approach utilized to summarize predictions across the bag: a parameter which can substantially influence the instances in which a model will succeed or fail. For instance, a max pooling-based approach considers the single sample most strongly associated with disease without incorporating any information from the bag's other samples. Meanwhile, a mean pooling of predictions within a bag might miss a disease diagnosis present in only a few samples.

 Recurrent neural networks, such as long short term memory (LSTM), excel in identifying relationships between correlated samples, such as in pattern recognition across time series data. Convolutional long short term memory (Conv-LSTM) \cite{convlstm} extends this capability to spatial data by making the operations of a LSTM convolutional. Conv-LSTM has been highly effective in characterizing changes in image patterns over time, such as video classification \cite{classify} and gesture recognition \cite{gesture}. Rather than identifying spatiotemporal patterns from time series image data, we propose the use of Conv-LSTM to “scan” through an imaging volume for disease presence without the need for expert annotations of diseased regions. In contrast to an MIL-based approach, our framework allows the detection of emphysema-associated image patterns on and between slices as it processes through the image volume. The network stores emphysema-associated image patterns across multiple bidirectional passes through a volume, and outputs a final set of features characterizing the entire volume without requiring a potentially reductive bag pooling operation. Our approach can make efficient use of weak, but readily accessible image labels (e.g. binary diagnosis of emphysema positive or negative) for abnormality detection within image volumes.

\vspace{-2mm}
\section{Methods}
\vspace{-4mm}
\subsection{Dataset and processing}
\vspace{-2mm}

A total of 8794 non-contrast CT volumes from 6648 unique participants enrolled in the National Lung Screening Trial (NLST) \cite{nlst_nejm,nlst_rad} were utilized. We used 3807 CT volumes from 2789 participants who were diagnosed with emphysema across the 3 years of study as positive samples and 4987 CT volumes from 3859 participants who were not diagnosed with emphysema in any of the 3 years as negative samples. 75\% of these scans, with balanced distribution of emphysema positive and negative patients, were utilized for model training. 4197 volumes from 3166 patients were used to directly learn model parameters, while 2434 volumes from 1319 patients were used to tune hyper-parameters and evaluate performance in order to choose the best-performing model. The remaining 2163 volumes (578 emphysema positive, 1585 emphysema negative), each from a unique patient, were held out for independent testing. Volumes were resized to 128x128x35, corresponding to an average slice {spacing} of 9 mm.  

\vspace{-2mm}
\subsection{Convolutional Long Short Term Memory (LSTM)}
\vspace{-2mm}

\begin{figure}[!htb]
    \centering
    \includegraphics[width=1\linewidth]{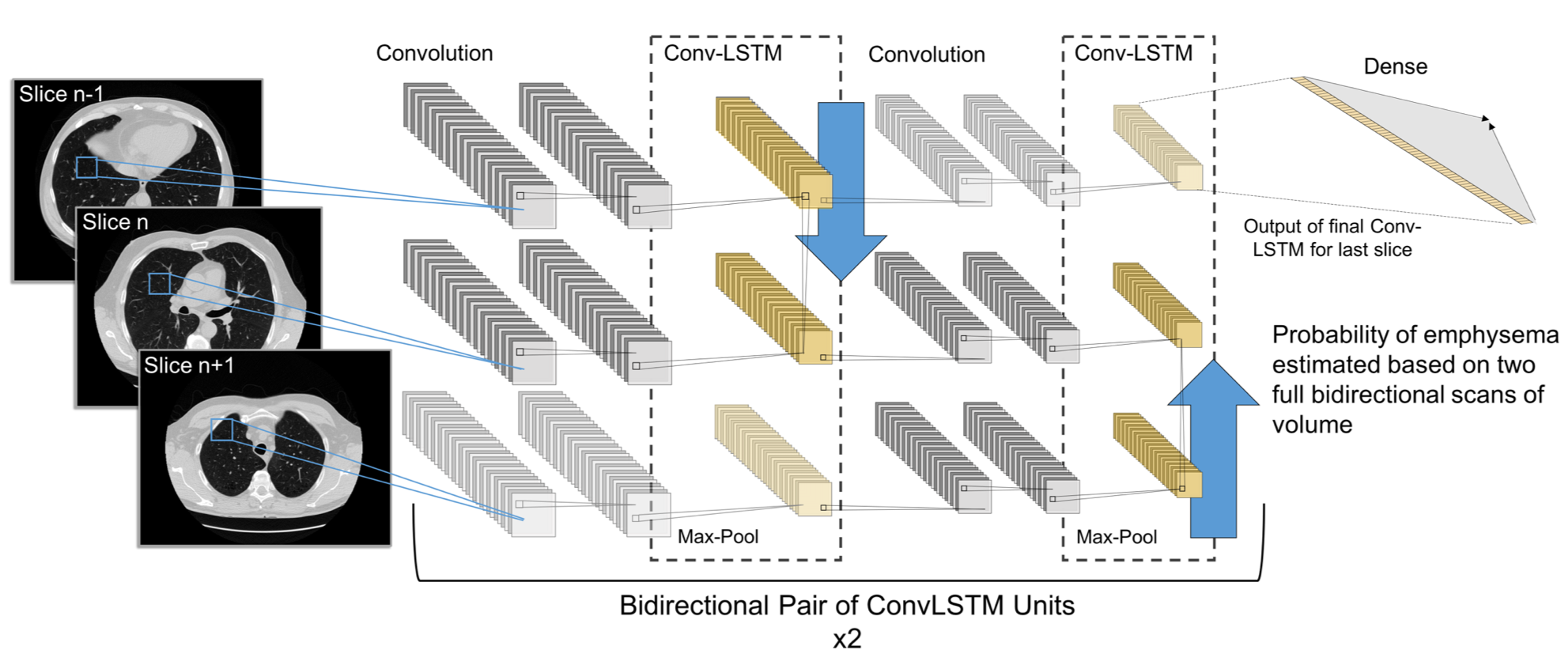}
    \caption{Conv-LSTM based neural network for emphysema detection from weakly labeled data.}
    \label{fig:arch}
\end{figure}

The architecture is comprised of four units each including convolution operations applied separately to each individual slice and a conv-LSTM to process the volume slice-by-slice. Two 3x3 convolutional layers with batch normalization are followed by max-pooling. The output from the convolutional layers for each slice are then processed sequentially in forward or reverse order by the conv-LSTM layer, which outputs a set of features obtained through convolutional operations with the input slice as well as previous slices within the volume. All layers within a unit share the same number of filters and process the volume in either ascending or descending order. The four convolutional units have the following dimensionality and directionality: Ascending 1: 32 filters, Descending 1: 32 filters, Ascending 2: 64 filters, Descending 2: 64 filters. The final Conv-LSTM layer outputs a single set of features, which thus summarizes the network's findings after processing through the full imaging volume multiple times. A fully-connected layer with sigmoid activation then computes probability of emphysema. The network, depicted in Fig. \ref{fig:arch}, comprises a total of 901,000 parameters. All models were trained for 50 epochs or until performance in the validation set ceased to improve. 

\subsection{Comparison Experiments}
\vspace{-2mm}

\paragraph{Multiple Instance Learning:} We implemented a MIL-based network where each slice of the CT volume was considered a sample from a bag. To this end, a purely convolutional network architecture resembling that of Fig. \ref{fig:arch}, with additional single-slice convolutional layers replacing conv-LSTM layers, to analyze each slice was implemented. A number of methods of summarizing predictions across the entire volume into a single bag probability were explored. The overall probability, $P$, for a bag containing $N$ samples with individual probability of emphysema, $p_i, i\in\{1,\cdots,N\}$, can be computed via the following approaches: 
\vspace{-2mm}
\begin{enumerate}
    \item{Max Pooling: } $P$ = max($p_i$)

    \item{Mean Pooling:} $P$ = $\frac{1}{N}$ $\sum^N_{i=1}$ $p_i$

    \item{Product Pooling:} $P$ = $1 - \prod^N_{i=1} 1 - p_i$

\end{enumerate}
\vspace{-3mm}

\paragraph{3D CNN:} Conv-LSTM was also compared with a 3D CNN similar to the structure of the 2D CNN with MIL, instead with a single dense layer and no pooling operation on the final convolutional layer. The number of kernels for each comparison model was increased to make its number of parameters relatively equivalent with our Conv-LSTM framework and ensure a fair comparison (Table \ref{table:results}).
\vspace{-2mm}

\section{Results and Conclusions}
\vspace{-4mm}

Convolutional-LSTM demonstrated strong performance in the identification of emphysema when trained using only weakly annotated imaging volumes, achieving an AUC=0.82. It outperformed a CNN with MIL regardless of pooling strategy (Max pooling: AUC=0.69, Mean Pooling: AUC=0.70, Product pooling: AUC=0.76). At the optimal operating point corresponding to the Youden Index \cite{youden}, our model achieved sensitivity and specificity of 0.77 and 0.74, respectively. Results for all evaluated models in the testing set are shown in Table \ref{table:results}. 

\begin{table}[!htb]
\centering
\begin{tabular}{|l|c|c|c|c|c|c|}
\hline
 & \textbf{Kernels} & \textbf{\# Parameters} & \textbf{AUC} & \textbf{Sensitivity} & \textbf{Specificity} & \textbf{F1} \\ \hline
\textbf{MIL - Max Pooling} & 64 & 1,011,393 & 0.69 & 0.59 & 0.68 & 0.63  \\ \hline
\textbf{MIL - Mean Pooling} & 64 & 1,011,393 & 0.70 & 0.76 & 0.57 & 0.66  \\ \hline
\textbf{MIL - Product Pooling} & 64 & 1,011,393 & 0.76 & 0.61 & 0.79 & 0.69 \\ \hline
\textbf{3D CNN} & 36 & 958,213 & 0.77 & 0.61 & 0.80 & 0.69 \\ \hline
\textbf{Conv-LSTM}& 32 & 901,793 & 0.83 & 0.77 & 0.74 & 0.75  \\ \hline

\end{tabular}
\caption{Emphysema detection results in the testing set (2,219 CT volumes) and model size.}
\label{table:results}
\end{table}
\vspace{-4mm}
Significantly, our approach requires no time-consuming annotation or manual processing of imaging data. Our framework allows training for disease detection from simple binary diagnostic labels, even if the disease is localized to only a fraction of the image. Therefore, our network can be readily trained from information easily attainable by mining radiology reports in an automated fashion. This capability significantly expands the pool of volumetric imaging data that can be practically utilized for such an application, and could allow easy retraining and finetuning of an algorithm when applied at a new hospital. Beyond emphysema, this approach is applicable to other disease/abnormality detection problems where the availability of volumetric imaging data exceeds the capacity of radiologists to provide manually delineated ground truth, but labels can be easily mined from radiology reports.   

\bibliographystyle{splncs}
{\renewcommand{\baselinestretch}{0.7}\bibliography{nipsmedimgbib}
}


\end{document}